\pdfoutput=1
\documentclass[11pt]{article}

\usepackage{EACL2023}
\usepackage{times}
\usepackage{latexsym}

\usepackage[T1]{fontenc}

\usepackage{graphicx}
\usepackage{amsmath}
\usepackage{bm}
\usepackage{subcaption}
\usepackage{caption}
\usepackage{multirow}
\usepackage[textsize=small]{todonotes}

\usepackage[utf8]{inputenc}

\usepackage{microtype}

\usepackage{inconsolata}

\title{Learning the Effects of Physical Actions in a Multi-modal Environment}

\author{Gautier Dagan, Frank Keller, Alex Lascarides  \\
  School of Informatics \\ 
  University of Edinburgh, UK \\
  \texttt{gautier.dagan@ed.ac.uk}, \texttt{\{keller, alex\}@inf.ed.ac.uk} \\
  }
\begin{document}
\maketitle

% TERMS:
% multi-modal not multimodal
% Pre-trained LM or Large Language Model (LLM)

\begin{abstract}
% NEW ABSTRACT:
Large Language Models (LLMs) handle physical commonsense information inadequately. As a result of being trained in a disembodied setting, LLMs often fail to predict an action's outcome in a given environment. However, predicting the effects of an action before it is executed is crucial in planning, where coherent sequences of actions are often needed to achieve a goal. Therefore, we introduce the multi-modal task of predicting the outcomes of actions solely from realistic sensory inputs (images and text). Next, we extend an LLM to model latent representations of objects to better predict action outcomes in an environment. We show that multi-modal models can capture physical commonsense when augmented with visual information. Finally, we evaluate our model's performance on novel actions and objects and find that combining modalities help models to generalize and learn physical commonsense reasoning better.
\end{abstract}

\section{Introduction}

Large Language Models (LLMs) are trained on large corpora of disembodied texts. 
They are typically pre-trained on a masked language modeling task: the model must predict a masked word in a text given its context.
LLMs have achieved state-of-the-art performance on many NLP tasks \cite{bert, lm}, but they can also fail on seemingly easy and obvious tasks and in unpredictable ways \cite{mccoy-etal-2020-berts, bommasani2021opportunities}.
Commonsense knowledge is shared knowledge and is often so obvious that it is absent from the LLMs' training data: people don't mention what is already known to their interlocutors. 
This includes physical commonsense information, including how executed actions affect the physical attributes of objects; e.g., shape and weight  \cite{Forbes2019DoNL}. 
Humans may learn such knowledge from their embodied environment. But LLMs, being trained on disembodied text, can make incorrect predictions about physical attributes and how these change when actions occur.
For instance, when asked what the weight of a 150 grams potato after it is sliced, GPT-3 \citep{lm} incorrectly answers 75 grams (see Appendix~\ref{appendix:gpt3} for the exact prompt).
GPT-3 is an LLM with 175 billion parameters, and nonetheless its disembodied existence limits its physical commonsense estimates.

\citet{zellers} inject physical commonsense information into LLMs via their model PIGLeT---a modified LLM that is trained on their PIGPeN simulated 3D environment dataset.
PIGLeT estimates how an environment changes as a result of specific actions. 
In training and testing, the model uses ground-truth symbolic representations of the environment but not the images: it ignores visual sensory observations. 
% PIGLeT does not use images.
These symbolic representations of objects in an environment are chosen to capture the possible effects of actions, and include attributes like \texttt{weight}, \texttt{size} and \texttt{temperature}.
However, in an embodied situation, an agent needs to use visual perception to estimate its interpretation of the scene.
Therefore, the symbolic representations should be treated as latent rather than observed.

\begin{figure*}[ht]
\centering
\includegraphics[width=0.75\textwidth]{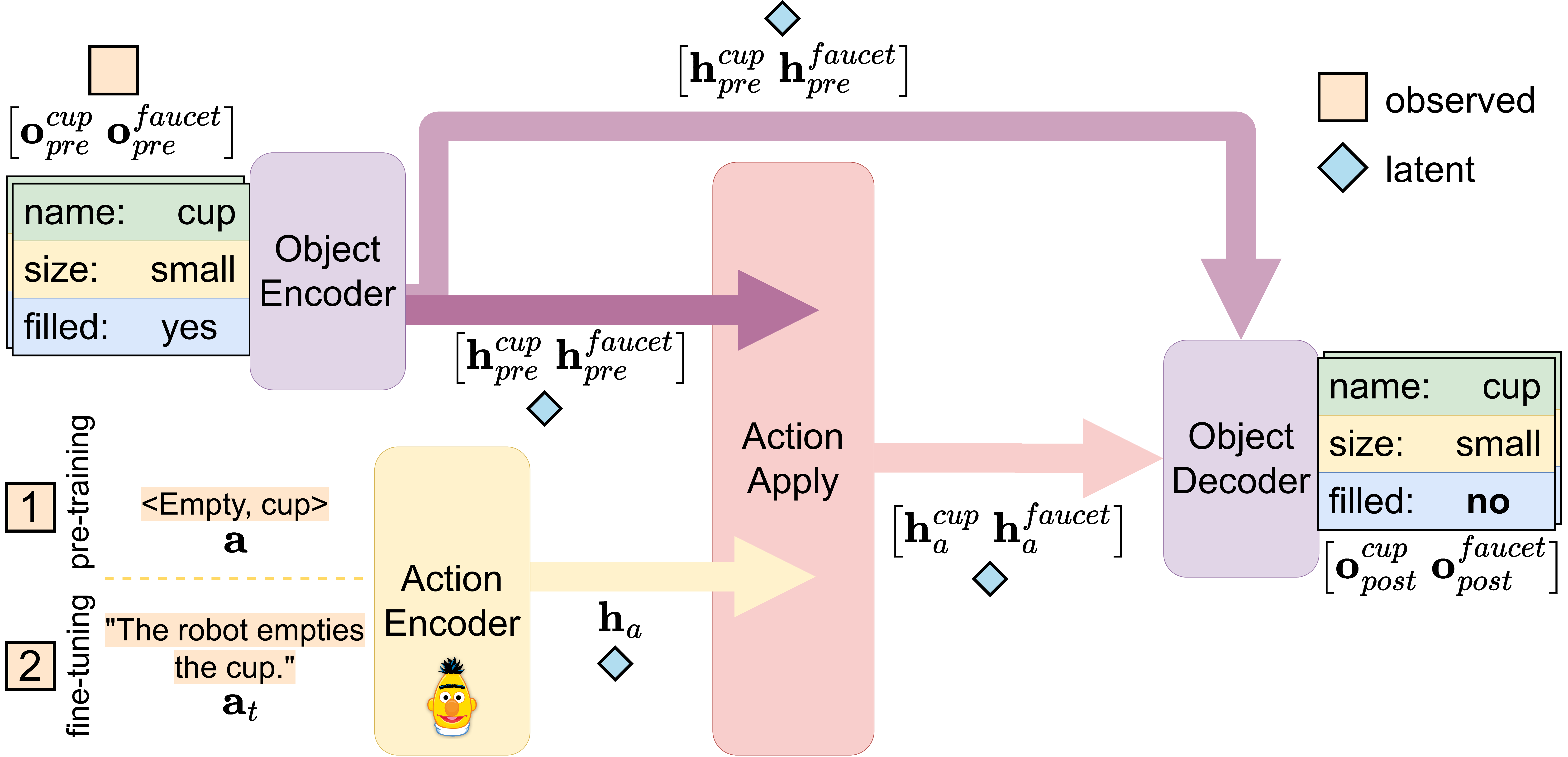}
\caption{\textbf{Original PIGLeT Physical Dynamics Model \cite{zellers}.} During pre-training the model receives as input the full symbolic representation of two objects ($\textbf{o}^{0}_{pre}$ and $\textbf{o}^{1}_{pre}$) before the action is taken and the symbolic representation of the action itself ($\textbf{a}$) and is tasked with predicting the attributes of the objects after the action ($\textbf{o}^{0}_{post}$ and $\textbf{o}^{1}_{post}$). 
During fine-tuning, the action encoder is replaced by an LLM to process a natural language description of the action being taken and with what objects.}
\label{fig:piglet}
\end{figure*}

We propose an alternative to the PIGLeT model, PIGLeT-Vis, which uses images directly as input into a multi-modal LLM to ground the model to its physical environment.
We compare our approach to the original PIGLeT model and evaluate the generalization capabilities gained from using image inputs.
At test time, our model foregoes symbolic labels: only the images and the name of the action are observed.
Thus our model tackles a more challenging task than the original PIGLeT model in that it must not only predict the effect of actions but also (indirectly) estimate the symbolic representations of objects in the images.
We also evaluate a model for predicting the effects of actions that trains on PIGPeN's images and their associated natural language (NL) descriptions, eliminating the need for formal symbolic representations. 

Our contributions are three-fold. First, we show that it is possible to predict the physical effects of actions from visual data. Second, we show that it is possible to learn the task on training data where formal symbolic representations, which are unobservable in real-world settings, are replaced with NL descriptions (which can be observed through natural interaction).
Third, we evaluate all our models in a stricter zero-shot setup to promote ways to train agents that generalize.
Overall our work paves the way for multi-modal models that learn the effects of actions in realistic environments.

% Our contributions are two-fold: (1) we filter a subset of PIGPeN that can be used to evaluate physical commonsense through vision, and we extend its zero-shot setup; and (2) we show that it is possible to develop physical commonsense reasoning from vision, paving the way for multi-modal models to learn the effects of actions in realistic environments.

\section{Related Work}

Commonsense reasoning has been highlighted as a potential weak point of LLMs in recent years \cite{generalisation_commonsense, Forbes2019DoNL, Bisk2020PIQARA}.
% Commonsense knowledge is typically shared knowledge and there are multiple types, such as knowing how the world reacts to change (physical) and how people respond to certain events (social).
Datasets such as PIGPeN \cite{zellers}, commonsenseQA \cite{commonqa}, VCR \cite{zellers_vcr} and GD-VCR \cite{yin2021broaden} help evaluate different aspects of commonsense reasoning in modern LLMs.
In this paper, we focus on physical commonsense reasoning, which involves understanding the (often) unexpressed rules of the physical world.

\citet{Forbes2019DoNL} reported that neural representations found it challenging to infer the link between actions and what they imply about the attributes of objects.
Accordingly, \citet{zellers_vcr} introduced the Visual Commonsense Reasoning (VCR) task to test how images can inform question answering models that tackle commonsense information. 
\citet{Bisk2020PIQARA} designed the PIQA benchmark to evaluate physical commonsense reasoning in LLMs through question answering.
\citet{sampat-etal-2021-clevr} proposed an extension to the CLEVR dataset, where an agent must reason and answer questions about a scene after a hypothetical action is taken.

Multiple approaches can improve the capabilities of LLMs in commonsense reasoning, such as using handcrafted knowledge graphs \cite{comet} or leveraging simulated environments \cite{zellers}.
PIGLeT, in particular, combines a traditional LLM and a ``Physical Dynamics'' model to ground an LLM \cite{zellers}.
The Physical Dynamics model enhances the commonsense knowledge of an LLM by fine-tuning it, using trajectories sampled from a realistic environment (see Figure~\ref{fig:piglet}).
Trajectories are an action and a pair of environment states (before and after the action) expressed in a formal symbolic representation.
\citet{zellers} found that fine-tuning LLMs with symbolic data from the simulated environment helped them outperform other models in physical commonsense reasoning tasks: in particular, predicting the effects of an action when executed in a particular state.

\begin{figure*}
\centering
\includegraphics[width=0.75\textwidth]{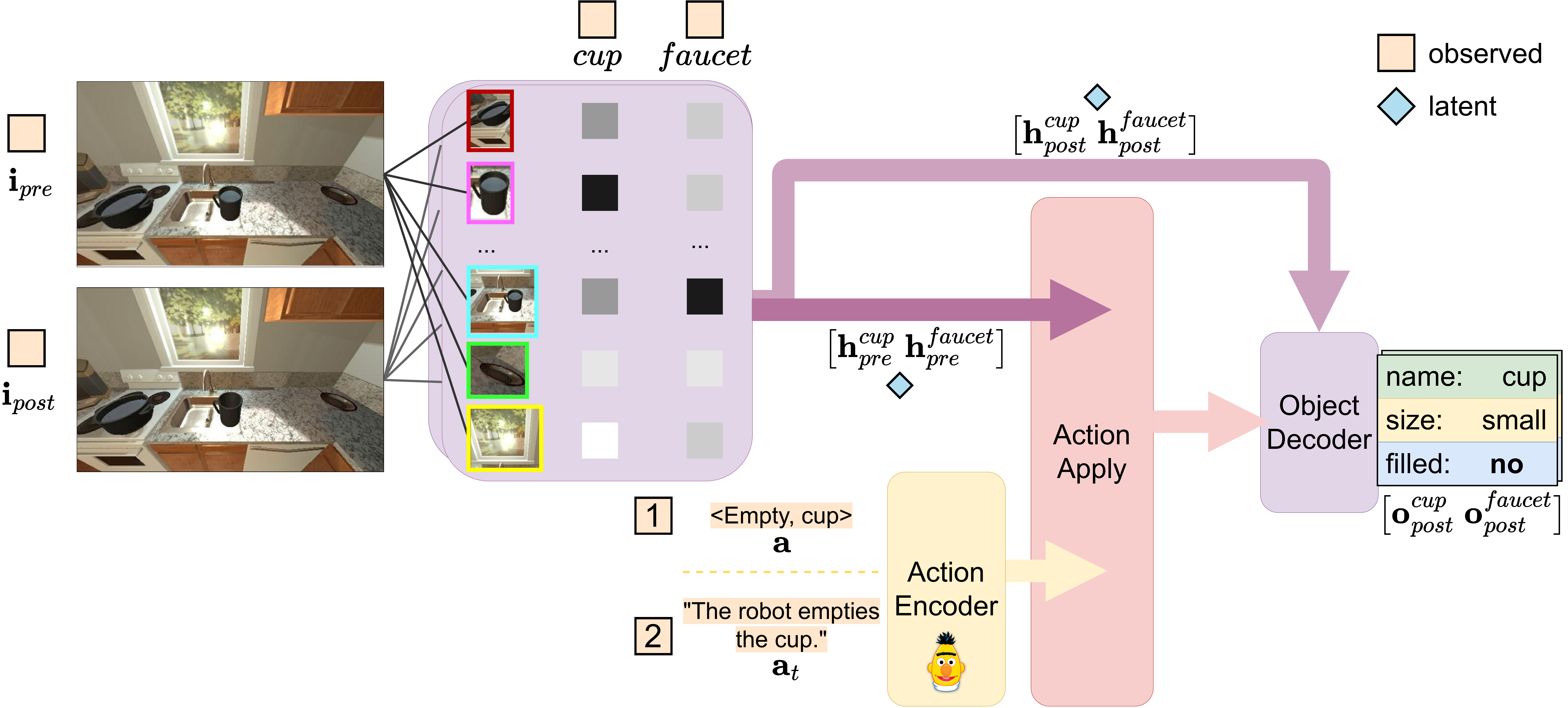}
\caption{\textbf{PIGLeT-Vis.} We introduce PIGLeT-Vis, where we modify the PIGLeT architecture to replace its Symbolic Object Encoder with a vision component that makes use of images of the environment before and after an action is taken to predict the symbolic representation of objects post-action. We use an attention mechanism over the extracted bounding boxes to obtain a visual hidden representation of an object given its name. The only remaining symbolic inputs during pre-training are the action description and object names.}
\label{fig:vision}
\end{figure*}

% multi-modal LLMs have also been proposed for the task of visual commonsense reasoning to incorporate images along with text.
Image inputs offer a way to ground an LLM, as they only require general alignment with a text or symbolic input and do not require the comprehensive environment ground-truth labels that PIGLeT uses.
% \citet{wang} were the first in formulating actions as transformations on sequential images.
\citet{gao-etal-2018-action} used multi-modal web data to learn actions and their effects from images and corresponding text descriptions.
\citet{zellers_vcr} used an off-the-shelf ResNet50 model \cite{resnet} to augment an existing BERT language model \cite{bert} with vision capabilities.
% CLIP \cite{clip} uses an alignment objective between images and text to train a model to map between the modalities.
Transformer models such as UNITER \cite{Chen2020UNITERUI}, ERNIE-ViL \cite{Yu2021ERNIEViLKE}, VisualBERT \cite{visualbert}, and ViLBert \cite{vilbert} have been applied to visual commonsense reasoning. 
These models use a joint transformer backbone for images and text and vary their pre-training objectives.
However, most of these models are trained on static text-image pairs: they aren't designed to capture the dynamics of an environment, particularly how object attributes change with actions.
Notably, recent work by \citet{hanna-etal-2022-act} uses CLIP \cite{clip} and MOCA \cite{moca} embeddings to predict a post-action image given a set of possible images.
In contrast, we focus on adapting an LLM with a vision-based component to predict the consequences of actions on the environment.

\section{Method}

We propose PIGLeT-Vis (Figure~\ref{fig:vision}) for learning the effects of actions on objects from images.
We use a pre-trained vision backbone, DETR \cite{detr}, as a Vision Object Encoder and combine it with a RoBERTa LLM \cite{roberta} as an Action Encoder.
We experiment with different configurations of inputs to measure the impact of the various components of our architecture.
In particular, we test a variation in which we remove the formal symbolic labels even in training, replacing them with NL text labels.
To evaluate our models, we use the PIGPeN dataset \cite{zellers}, which consists of a symbolic and visual representation of an environment before and after an action is taken.
However, we filter PIGPeN to create a viable testing ground for visual grounding of physical actions and more accurately measure generalization capabilities of models.

\subsection{Architecture}

PIGLeT-Vis (shown in Figure~\ref{fig:vision}) consists of separate components, which can combine multi-modal inputs in different ways. 
Through this modular approach, we can turn off specific components to evaluate how different inputs and model structures affect performance on the task.
We test models with and without symbolic inputs and image inputs.
% Apart from our Vision Encoder, the architecture of other components is kept the same as the PIGLeT model.
For all components, we use a dropout of $p=0.1$ in between layers and a default hidden layer size of $h=64$.

\subsubsection{Object Encoder}

We reproduce \citet{zellers}, where all actions are assumed to involve two objects, $\textbf{o}^0$ and $\textbf{o}^1$, and the symbolic representation of objects are encoded in an Object Encoder model.
The symbolic representation of an object before the action is represented by $\textbf{o}_{pre}$.
Both objects ($\textbf{o}^{0}_{pre}$ and $\textbf{o}^{1}_{pre}$) in the environment are described by a vector of 38 attributes, chosen on the basis that they are the kinds of physical attributes that are influenced by actions.
They describe an object as small/large, cold/hot, empty/full, etc. 

We first embed these symbolic object attributes using an embedding layer $\textbf{E}^{e \times h}$, where $e=329$ is the total number of unique attributes and $h$ is our hidden size.
For an object $k$:
\begin{align}
    \hat{\textbf{o}}^k_{pre} &= \textbf{E}(\textbf{o}^k_{pre})
\end{align}
%
% As a result of our embedding, we obtain $\hat{\textbf{o}}$ of dimensions $2 \times 38 \times h$ from our $\textbf{o}_{pre}$ of dimension $2 \times 38$.
%
The Object Encoder $\textbf{O}_{encoder}$ takes in the embedded object attributes through a set of multi-head attention layers to encode the symbolic representation of each object.
We use the default Pytorch implementation of the Transformer Encoder \cite{pytorch} with three layers and 4 heads.
The first encoded output of each object sequence is used for representing the entire object.

\begin{align}
    \textbf{h}^k_{pre} &= \textbf{O}_{encoder}(\hat{\textbf{o}}^k_{pre})
\end{align}

\subsubsection{Action Encoder}

Actions are encoded either as a symbolic triplet $\langle$action, action object, action receptacle$\rangle$ or as an annotated text describing an action being taken (e.g., ``robot empties the cup'').

During pre-training, the Action Encoder $\textbf{A}_{pretrain}$ uses an action embedding layer $\textbf{E}'$ to embed the first dimension of the action, and re-uses the object embedding layer $\textbf{E}$ to embed the action object name $a_o$ and action receptacle name $a_r$.
The action embedding layer $\textbf{E}'$ has dimensionality $10 \times h$ for the 10 distinct actions.
The three embedded representations are summed and passed to the Action Encoder's linear layers to produce $\textbf{h}_a$ (see equation \ref{eq:ha}). 
Similarly to \citet{zellers}, a \texttt{tanh} activation is applied after each linear layer.

\begin{align}
\label{eq:ha}
    \textbf{h}_a &= \textbf{A}_{pretrain}(\textbf{E}'(\textbf{a}) + \textbf{E}(a_o) + \textbf{E}(a_r)))
\end{align}

When fine-tuning on the annotated dataset, the action input is text and therefore we switch out the Action Encoder $\textbf{A}_{pretrain}$ for $\textbf{A}_{\mathit{finetune}}$---our text-based Action Encoder.  $\textbf{A}_{\mathit{finetune}}$ uses a RoBERTa-base\footnote{\label{hugface}Implementation and pre-trained model weights are taken from the Huggingface library \cite{huggingface}.} model \cite{roberta} to process a tokenized version of the text input $\textbf{a}_t$. 
The first token ([\texttt{CLS}]) of the RoBERTa output layer is used to represent the action sequence and then passed through a linear layer to map the dimensionality of the hidden states from $256$ to $h$. 

\begin{align}
    \textbf{h}_a &= \textbf{A}_{\mathit{finetune}}(\textbf{a}_t)
\end{align}

\subsubsection{Vision Object Encoder}

The Vision Object Encoder takes in images ($\textbf{i}_{pre}$ and $\textbf{i}_{post}$) to provide a visual representation of each object $k$ before and after ($\textbf{h}^k_{pre}$ and $\textbf{h}^k_{post}$).
We use the DETR\textsuperscript{\ref{hugface}} \cite{detr} model as a backbone to predict $N$ bounding boxes in a pair of images (pre- and post-action).
As DETR is pre-trained on the COCO object detection dataset \cite{coco}, its predicted object labels do not align with those in PIGPeN.
Therefore, we instead learn a mapping between the predicted bounding box representations and the PIGPeN objects.
For each image, we obtain a hidden representation $\textbf{h}_b$ of dimensionality $N \times 256$ where $N=100$.

We use an attention mechanism over the bounding boxes' hidden representation, conditioned on the object names.
For a given object $o^{k}$, its conditional representation $\textbf{h}_c^k$ is the encoded name of the object: $\textbf{E}(o^k_{name})$.
We can therefore obtain the attention score of a given object $o^k$ and image $\textbf{i}_{m}$ by calculating the alignment between the conditional representation $\textbf{h}_{c}^{k}$ and the hidden representations of bounding boxes $\textbf{h}_{b_{m}}$:
\begin{align}
    \textbf{h}_{b_{m}} &= \text{DETR}(\textbf{i}_{m}) \\
    \bm\alpha_{m}^{k} &= \text{Softmax}\left(\sum^h_{i=1}  (\textbf{h}_{c}^{k}  \textbf{h}_{b_{m}})_i \right)
\end{align}
We obtain the final representation for a given object and image by multiplying our attention scores $\bm\alpha$ with the extracted output representation from DETR and summing along the bounding box axis:
\begin{align}
    \textbf{h}^k_{o_m} &= \textbf{W} \left( \sum^b_{j=1} (\bm\alpha_{m}^{k} \textbf{h}_{b_{m}})_j \right)
\end{align}
We use a final output layer $\textbf{W}$ to decrease the dimensionality of $\textbf{h}_o$ from the DETR dimensionality of $256$ to $h$.

Through the Vision Object Encoder, we replace the previously symbolic inputs with images and can extract $[ \textbf{h}^0_{pre}  \textbf{h}^1_{pre} ]$ and $[\textbf{h}^0_{post}  \textbf{h}^1_{post}]$ from $\textbf{i}_{pre}$ and $\textbf{i}_{post}$ respectively.
Note that we make the implicit assumption that $\textbf{i}_{pre}$ and $\textbf{i}_{post}$ contain the information necessary to predict object attributes of the objects post-action.

\subsubsection{Action Apply}

The Action Apply Model $\bm{\beta}$ is a simple fuse operation (concatenation in the hidden dimension) followed by three linear layers, which combine the action representation $\textbf{h}_a$ and an object representation of the scene pre-action $\textbf{h}^k_{pre}$.
The model outputs an object's representation $\textbf{h}^k_{a}$, containing information conditioned all inputs:
\begin{align}
    \textbf{h}^k_{a} &= \bm{\beta}(\textbf{h}_a, \textbf{h}^k_{pre})
\end{align}

\subsection{Object Decoder}

Finally, the Object Decoder is a transformer module that maps the object representations $h_o$ from the pre-action state back to 38 symbolic attributes.
It uses a default three layer Transformer Decoder \cite{pytorch} that takes the hidden representation from the Action Apply $\textbf{h}^k_{a}$ as an encoded memory state and $\textbf{h}^k_{pre}$ as the source sequence to predicts a label for each attribute.
\begin{align}
    \dot{\textbf{o}}_{post}^k &= \textbf{O}_{decoder}(\textbf{h}^k_{a}, \textbf{h}^k_{pre})
\end{align}
When we use image inputs, we also have access to the post-action visual representation and can therefore use $\textbf{h}^k_{pre}+\textbf{h}^k_{post}$ instead of $\textbf{h}^k_{pre}$.

The output has post-action object states $\dot{\textbf{o}}^k_{post}$ which are compared to the ground truth $\textbf{o}^k_{post}$ to calculate cross-entropy. 
As an additional loss, we also use the cross-entropy between $\dot{\textbf{o}}^k_{pre}$ and $\textbf{o}^k_{pre}$ by passing an empty $\textbf{h}^k_{a}$ to force the Object Decoder to recreate the attributes in the pre-action state.
We weight both losses equally.

% \section{Experimental Setup}
\subsection{Evaluation Metrics}

Since our task involves predicting 38 attributes for two different objects per example, we follow \citet{zellers} and report different types of accuracy metrics on the test set (after fine-tuning).
We measure the overall accuracy by scoring how many objects have all attributes correctly predicted (exact match).
Note that this is a high bar for a model where the symbolic representations are latent: to predict an object correctly, our model must first estimate its attributes before the action and then estimate whether and how these change given an action.
So we also measure the attribute-level and action-level accuracies of each model, so as to explore which attributes and actions are more difficult to predict than others.

\subsection{PIGPeN-Vis Dataset Split}
\label{sec:corpus-split}

To evaluate physical commonsense reasoning using PIGLeT-Vis, we filter PIGPeN \cite{zellers} to create a subset (PIGPeN-Vis) which we use for all our experiments.
We motivate PIGPeN-Vis as a way to isolate the effects of adding our vision component, because while PIGPeN already has images, these images were not used in PIGLeT.

The PIGPeN dataset consists of trajectories of an environment before ($pre$) and after ($post$) an action is taken.
Each trajectory contains representations of two distinct objects before and after. 
One of the objects is usually targeted by the action, while the other acts as a distractor.
In addition, image pairs $(\textbf{i}_{pre}, \textbf{i}_{post})$ for each trajectory are provided, where each image is snapshot of the simulated photo-realistic 3D environment which contains the objects in view (see Appendix~\ref{appendix:pigpen} for an example). 
Each image is an RGB image of dimensions $640 \times 385$.

The original dataset is separated into two distinct sets:

\begin{enumerate}
    \item A \textit{pre-training} set of $278,009$ trajectories, which includes the symbolic representations of objects $\textbf{o}$ before and after a symbolic action $\textbf{a}$ is taken. A separate validation set of $33,042$ examples is also included.
    \item A \textit{fine-tuning} set of $1,000$ trajectories which has been annotated to replace the symbolic action $\textbf{a}$ with a textual representation $\textbf{a}_{\textbf{t}}$ describing the action. Separate validation and test sets of $500$ examples each are also included. All metrics are reported on the test set.
\end{enumerate}

In PIGPeN, the object states $\textbf{o}_{pre}$ and $\textbf{o}_{post}$ contained $40$ different attributes and $13$ different actions $\textbf{a}$.
Attributes range from intrinsic such as \texttt{name} or \texttt{moveable} to stateful such as \texttt{distance} or \texttt{isCooked}.
In forming PIGPeN-Vis, we remove two attributes and three actions from the dataset to obtain $38$ attributes and $10$ possible actions (see Appendix~\ref{appendix:pigpen} for more details).

\subsubsection{Viewpoint and Action Filtering}
\label{sec:viewpoint_filtering}

Since the PIGPeN images were not generated with the goal of being used as input data, we identified several issues with the quality of certain scenes.
A notable difficulty is that in some cases, the before and after images are not captured from the same camera angle or they have different lighting conditions.
Changing orientations and lighting conditions makes it difficult to use an image pair $(\textbf{i}_{pre}, \textbf{i}_{post})$ to isolate the outcome of an action.
Conversely, image pairs with too few perceivable differences also break our assumption that the changes in the environment are perceivable.
Therefore, we filter the dataset using pixel statistics to remove image pairs that have either large perceivable differences (likely due to changes in viewpoint) or small perceivable differences (where the action's results are not visually salient enough) (see Appendix~\ref{sec:filter_stats}).
We exclude $~15.4\%$ of the total dataset through visual filtering of the original dataset.

\subsubsection{Zero-Shot Filtering}

To evaluate the generalization capabilities gained from a vision component, we further filter the dataset to exclude a subset of training examples.
Unlike the original PIGPeN dataset which only tested for zero-shot generalization at the level of the fine-tuning data, we remove all instances with selected specific objects or action-object pairs from all training and validation sets.
To minimize the effect of removing examples from the dataset, we pick objects and action-object pairs with an already low number of samples in the training sets.
In total, we exclude 14 objects and 27 action-object pairs, which amounts to less than $3\%$ ($6,816$ samples) of the remaining training sets (see Appendix~\ref{sec:zero_app}).
These zero-shot examples comprise around~$10\%$ of the test set.

After both filtering stages, PIGPeN-Vis contains a pre-training dataset of $232,625$ trajectories with a validation set of $26,823$, and a fine-tuning training set of $750$ examples with a validation set of $367$ examples and a test set of $398$ examples.

\subsection{Training Configurations}

We evaluate the impact of the vision component on PIGPeN-Vis through five different setups:
\begin{itemize}
    \item \texttt{base}: We implement a baseline model without symbolic object inputs. Our implementation removes the Object Encoder entirely, such that the model must predict the attributes of objects solely from knowing the action and the object names that it relates to. This model acts as a lower bound on the capabilities of the vision model: its performance would match the vision model if images are irrelevant to solving the task.
    \item \texttt{\texttt{base+symbolic}}: This is our implementation of the original \citet{zellers} PIGLeT model, shown in Figure~\ref{fig:piglet}. This model acts as an upper bound on the capabilities of the vision model since it observes the true symbolic representations of objects before the action (which the vision model must estimate).
    \item \texttt{base+images}: This is our proposed PIGLeT-Vis, shown in Figure~\ref{fig:vision}, where the Vision Object Encoder replaces the previously symbolic Object Encoder. This model leverages the before and after images of the environment as well as the name of the objects to extract representations of the object attributes.
    \item \texttt{base+symbolic+images}: We sum the hidden symbolic representations of objects with their visual representations in a unified model. Through this setup, we evaluate whether images can provide additional information to the already comprehensive symbolic representations.
    \item \texttt{base+images+text-labels}: We convert the symbolic representations of the labels for the object names and actions to their text label and encode them using a frozen LLM during pre-training. We use the same LLM to encode the text labels that we later use in the fine-tuning stage. This setup replaces all symbolic inputs from the pre-training stage to only language and image inputs.
\end{itemize}
Note that there are a few differences between the original \citet{zellers} model and our implementation of \texttt{base+symbolic}.
For instance, for simplicity, we opted to use an off-the-shelf RoBERTa-base \cite{roberta} model instead of training our own custom GPT2 \cite{radford}.
Additionally, we also reduce the dimensionality of the PIGLeT layers from $h=256$ to $h=64$. 
We found that not only does this allow faster training times as it shrinks the Physical Dynamics model from $11.9$ million parameters to $2$ million parameters, it also improves the overall accuracy by a small margin ($+1.51\%$).

We train each model for $80$ epochs with a batch size of $256$ using the Pytorch implementation of the Adam optimizer \cite{adam} and a learning rate of $10^{-3}$ during pre-training and $10^{-5}$ during fine-tuning. 
We run each setup over $10$ different seeds and report the average and standard deviation for each metric (see Appendix~\ref{appendix:training_details} for more details).

\begin{table}[t]
\centering
\scalebox{0.72}{
\begin{tabular}{lll}
\hline
&\multicolumn{2}{c}{\textbf{Accuracy} ($\% \pm \sigma$)}\\
&Overall & Zero-Shot\\
\hline 
\texttt{base}& $21.23\pm0.72$& $5.34\pm2.77$\\
\hline 
\texttt{base+symbolic (PIGLeT)}& $85.03\pm0.45$& $39.04\pm3.37$\\
\texttt{base+symbolic+images}& $86.01\pm0.89$& $35.89\pm3.47$\\
\hline 
\texttt{base+images (PIGLeT-Vis)}& $45.47\pm1.50$& $7.53\pm2.60$\\
\texttt{base+images+text-labels}& $47.55\pm2.10$& $8.90\pm3.24$\\
\hline
\end{tabular}
}
\caption{Overall and zero-shot accuracies (PIGPeN-Vis)}
\label{tab:results}
\end{table}

\begin{table*}[t]
\centering
\small
\begin{tabular}{lc@{~~}c@{~~}c@{~~}clc@{~~}c@{~~}c@{}}
\hline
& \multicolumn{4}{c}{\textbf{Action Accuracy} (\%)}&&\multicolumn{3}{c}{\textbf{Attribute Accuracy} (\%)} \\
&\texttt{Open}&\texttt{Pickup}&\texttt{ToggleOn}&\texttt{Slice}&&\texttt{size}&\texttt{distance}&\texttt{temperature}\\
\hline
\texttt{base}& $8.33$& $10.96$& $27.38$& $22.13$&& $73.78$& $51.01$& $95.91$\\
\hline
\texttt{base+symbolic (PIGLeT)}& $85.73$& $80.48$& $\bm{96.90}$& $75.41$&& $94.98$& $95.13$& $\bm{99.85}$\\
\texttt{base+symbolic+images}& $\bm{88.75}$& $\bm{86.14}$& $92.86$& $\bm{81.31}$&& $\bm{96.35}$& $\bm{96.13}$& $99.59$\\
\hline
\texttt{base+images (PIGLeT-Vis)}& $20.83$& $33.49$& $70.24$& $41.64$&& $87.03$& $76.62$& $96.10$\\
\texttt{base+images+text-labels}& $22.92$& $40.12$& $67.14$& $45.57$&& $87.89$& $78.06$& $96.72$\\
\hline
\end{tabular}
\caption{Action and attribute specific accuracies for a subset of actions and attributes; for a comprehensive table with standard deviations see Appendix~\ref{appendix:results}. \texttt{size} and \texttt{distance} each have eight possible classes while \texttt{temperature} has three.}
\label{tab:results-act-att}
\end{table*}

\section{Results and Discussion}

We evaluate all models on our PIGPeN-Vis split and report the overall (exact match), zero-shot, action-level, and attribute-level accuracy results for all setups in Tables~\ref{tab:results} and~\ref{tab:results-act-att}. 
For completeness, we also evaluate models on the original PIGPeN to contrast the effects of our filtering operations (see \S\ref{sec:corpus-split} and Appendix~\ref{appendix:results}) and find PIGPeN-Vis is a more challenging subset for all models.

The \texttt{base} model provides a low bar estimate of what is achievable using only the action encoder inputs.
Unsurprisingly, the \texttt{base} model performs worst on overall accuracy, which demands an exact match of all attributes.
It does relatively well on (individual) attribute-level accuracy, primarily because it predicts the most common attribute for each object.
Some actions are also easier than others---for instance, the model reaches $27.38\%$ accuracy on \texttt{ToggleOn} from only knowing the action and object names.
This is likely because \texttt{ToggleOn} is constrained to a small set of objects and effects.

% We also note that for certain attributes such as \texttt{isBroken}, the accuracy for all models is $>99\%$ as only certain objects can be broken and none of the 10 actions modify this attribute, therefore it acts almost as an intrinsic attribute and is thus much easier to learn.

\begin{figure*}[ht]
    \centering
     \includegraphics[width=\textwidth]{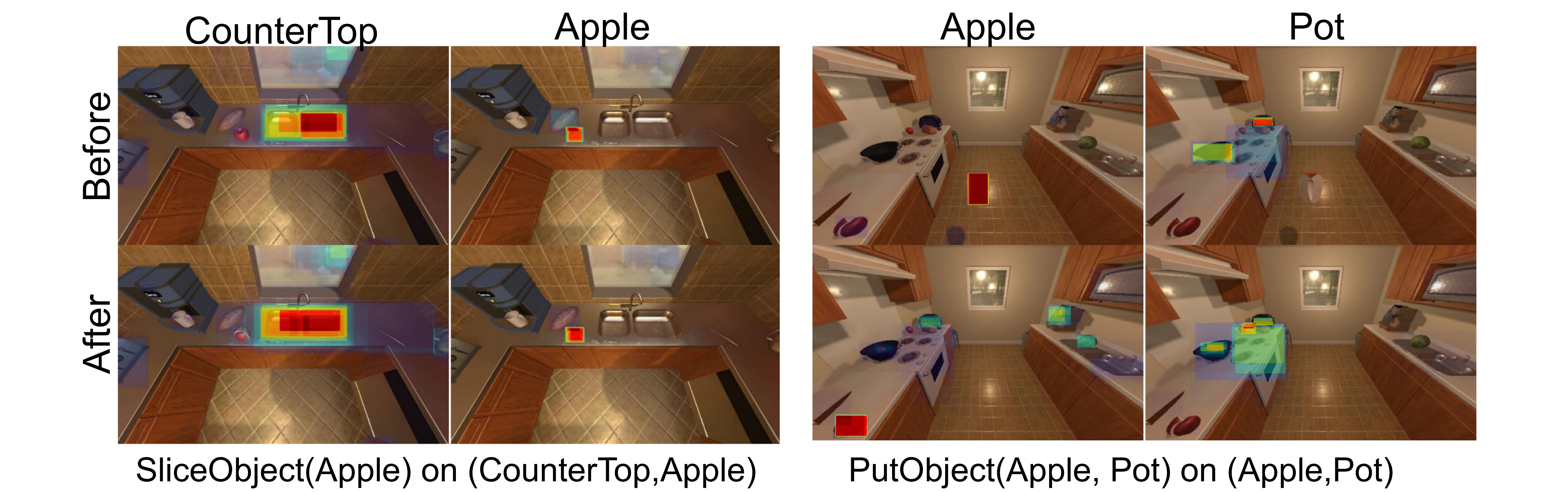}
\caption{We visualize the attention of the Vision Object Encoder from a trained \texttt{base+images} model on two different actions and environments. The left grid focuses on the effect of \texttt{Slice(Apple)} on \texttt{CounterTop} and \texttt{Apple}, while the right grid focuses on the effects of \texttt{Slice(Apple)} on  \texttt{Apple} and \texttt{Pot} objects.}
\label{fig:attention}
\end{figure*}

Our \texttt{base+symbolic} model obtains similar results to the original implementation by \citet{zellers}, with an overall accuracy of $85.03\%$.
However, it performs much worse on the zero-shot split ($39.04\%$) than the original PIGLeT model reported ($80.2\%$) \cite{zellers}.
This disparity can be explained by the fact that the original zero-shot PIGPeN dataset was not a true zero-shot dataset, because the Physical Dynamics model was exposed to the ``unseen'' objects in its pre-training.
The \texttt{base+symbolic} model provides a high bar estimate of what could be achievable if: (i)~$\textbf{i}_{pre}$ and $\textbf{i}_{post}$ capture the symbolic environment; and (ii)~the Vision Object Encoder can subsequently extract these features.
However, as we will argue in Section~\ref{sec:limitations}, both (i) and (ii) are unrealistic given the constraints of both the dataset and the model.

Our \texttt{base+images} (PIGLeT-Vis) model scores $45.28\%$ in overall accuracy but only $7.53\%$ on the zero-shot set.  
Nevertheless, it outperforms the \texttt{base} model in overall accuracy ($p<0.0001$) and in zero-shot accuracy ($p = 0.08$), which demonstrates that the images improve the prediction of the effects of actions.
The \texttt{base+images} model also performs significantly better than \texttt{base} on difficult attribute-level accuracies such as \texttt{distance} ($p<0.0001$).
However, as before, accuracy on individual attributes benefits from the skewed distributions of their values and does not necessarily translate to high scores on predicting all 38 attributes correctly.

Utilizing both images and symbolic representations as inputs helps the \texttt{base+symbolic+images} model outperform purely symbolic inputs in overall accuracy, from $85.03\%$ to $86.01\%$ ($p < 0.01$).
However, image inputs also decrease the model's zero-shot performance from $39.04\%$ to $35.89\%$, although this isn't statistically significant ($p = 0.05$) due to high variance.
We suspect that this high variance is caused by an increase in noise in the model resulting from adding images to the symbolic model. 
However, the overall picture is more complicated, as images can also provide gains on certain actions (e.g.,~\texttt{PickUp} accuracy increases from $80.48\%$ to $86.14\%$) even though it causes a decrease in many other cases (e.g.,~\texttt{ToggleOn}).

Finally, when we utilize NL descriptions to replace the formal symbolic inputs (action name and object names), \texttt{base+images+text-labels} improves overall accuracy when compared to \texttt{base+images} from $45.47\%$ to $47.55\%$ ($p = 0.02$).
Text inputs appear to improve zero-shot accuracy, but not by a statistically significant margin ($p = 0.31$).
Accuracy also improves in most actions, for instance the \texttt{Slice} accuracy improves from $41.64\%$ to $45.57\%$ ($p = 0.03$).
So the NL descriptions inform the task in a beneficial way, over and above the raw images.  
But encoding the labels as text rather than formal symbolic representations also adds noise.

Nevertheless, text labels improve accuracy on actions where the semantic information contained in the label provides a richer context to help generalize to similar objects.
For instance, a ``cup'' and a ``mug'' are semantically close, and thus learning the effects of actions on a ``cup'' might help the model predict the same effects on a ``mug'' even if the word forms are different.
% In contrast, the formal symbolic representations treat the predicate symbols \texttt{cup} and \texttt{mug} as unrelated, and so don't benefit from the lexical relationships that the RoBERTa-base LLM captures, which form a part of the encoding algorithm for NL descriptions in our task.
In contrast, the formal symbolic representations treat the predicate symbols \texttt{cup} and \texttt{mug} as unrelated, and so don't benefit from the lexical relationships that the LLM captures.
Fully removing the symbolic representations allows us to adapt our model to any possible unseen object during test time.
\texttt{base+images+text-labels} is adaptable to general settings without knowing the symbolic mapping of objects and actions in the environment.

The results of both \texttt{base+symbolic+images} and \texttt{base+images+text-labels} make the case multi-modal modeling of commonsense reasoning, as both language and images are complementary to generalize to unseen settings.

\subsection{Qualitative Attention Maps}

Visualizing attention is another benefit of a vision component, as we can see what the model focuses on and partially explain its predictions. Figure~\ref{fig:attention} shows two separate examples and corresponding attention maps. In the left example, \texttt{base+images} is tasked with predicting the attributes of \texttt{CounterTop} and \texttt{Apple} after the \texttt{Slice} action is applied on the \texttt{Apple}. In the right example, the \texttt{Put} action is applied on the \texttt{Apple}, and the model must predict the attributes of the \texttt{Apple} and the distractor object \texttt{Pot}. The two rows are the before and after images ($\textbf{i}_{pre}$ and $\textbf{i}_{post}$), and the two columns are the two objects used to condition the attention. 
The attention maps display the strength of the attention for each bounding box given an object name.

Both examples in Figure~\ref{fig:attention} show that the Vision Object Encoder can map known objects to relevant bounding boxes. 
The model successfully tracks the \texttt{Apple} in both cases by placing the most weight on the bounding box targeting the \texttt{Apple}.
However, these examples also show the difficulty of this task---the environments are realistic and can be filled with more than one instance of an object.

\section{Conclusion}

In this paper, we tackle the task of predicting the effects of actions on objects' physical attributes.
In contrast to \cite{zellers}, our model does not treat the formal symbolic representation of the images as observed.
Instead, PIGLeT-Vis supports inference when the inputs are images alone or images plus NL descriptions and a phrase denoting the action (e.g.,~``the robot empties the cup''). 
While PIGPeN offers challenges for applying a multi-modal approach, our model can extract useful information from images, opening the door for generalizing learning physical commonsense to real-world data.
Importantly, our PIGPeN-Vis split can be used to evaluate the zero-shot capabilities of different model configurations.
% Zero-shot evaluation is crucial as it indicates how well these models could perform after being deployed.
Moreover, while \texttt{base+symbolic} still outperforms \texttt{base+images}, it does so without estimating the attributes of objects and thus solves a much easier but unrealistic task.
Through \texttt{base+images+text-labels}, we show that, when replacing symbolic inputs, the best solution is to complement image inputs with NL descriptions to leverage information from both modalities.
Finally, our results show the need to improve the generalization capabilities of multi-modal models such that they can learn and adapt to unseen situations.

% In future work, we will explore an extension to the \texttt{base+images} model to entirely remove the need for the symbolic representation of the object name in the attention mechanism of the Vision Object Encoder. 
% Instead, we hope to replace its input by an LLM representation that makes use of the text label of the object instead.
% This would take advantage of the LLMs' pre-training regime and could improve the zero-shot performance of the \texttt{base+images} model.

% In future work, we will consider testing physical commonsense reasoning on datasets which include the segmentation map of the environment for a given image, such that the image encoder can be jointly fine-tuned with the LLM.
% This would also allow us to disentangle the attention mechanism when more than one instance of an object is present in an image.
% Furthermore, PIGPeN also suffers from arbitrary choices, such as which objects in the environment are distractors and what attributes to track.
% Attributes such as \texttt{isBreakable} depend on contexts, and the simulation's choice of setting and objects might teach the model wrong truths about the world.
% Future work could explore better ways to objectively capture the physicality of the world and prefer picking specific attributes over others depending on how important they might be to human perception.

\section{Limitations}
\label{sec:limitations}

There are several limitations to our approach that result directly from the inherent limitations of PIGPeN and our proposed Vision Object Encoder respectively. 

% \subsection{Limitations of PIGPeN}
% \label{sec:pigpen_limits}

PIGPeN was not originally designed for testing commonsense reasoning using images and contains numerous inconsistencies which cannot all be solved with the PIGPeN-Vis split obtained from filtering (Section~\ref{sec:viewpoint_filtering}).
Given the presence of non-physically salient attributes such as \texttt{temperature}, images are not guaranteed to fully capture their symbolic representations. 
PIGPeN includes certain attributes which are not discernible from images, e.g., even humans would be unable to tell a hot plate from a cold plate from vision alone.
The images in PIGPeN can also contain more than one object (e.g.,~more than one cup) without ever specifying which one the symbolic representation refers to.
This causes difficulty for our approach because judging specific attributes such as distance is impossible if there are two cups at different distances from the viewpoint.
Additionally, PIGPeN also discretizes continuous variables such as  \texttt{distance} into categories which can be hard to disambiguate.

To approach the accuracy of \texttt{base+symbolic} with our vision component, we also need a vision representation from which to correctly estimate all latent attributes.
Even if images are assumed to be perfect representations of the symbolic environment, the model still has to extract each of the 38 attributes correctly for both objects using only two images.
% The difficulty of the task is highlighted by the zero-shot setting where \texttt{base+images} only obtains $9.28\%$.
% The extraction of attributes fails when the attention mechanism within the Vision Object Encoder, which has been trained to map bounding boxes to object names, is shown unseen objects.
% Therefore the impressive zero-shot capabilities of \texttt{base+symbolic} are misleading, and can in part be explained by the fact that the model still receives the symbolic representation of the unseen objects before the action is taken.
% \texttt{base+symbolic} only has to predict the effect of an action on certain attributes and can copy the unchanged attributes in its outputs.
% Whereas \texttt{base+images} faces the much harder task of not only predicting the effects of an action but also predicting the attributes of completely unseen objects to begin with.
It is possible (and likely) for the vision detection backbone to miss the target object entirely because it is not trained to detect the specific object in question. 
We see this effect in Figure~\ref{fig:attention}, where the model falls back to using a bounding box around the sink area to describe the \texttt{CounterTop} object.
The DETR vision model used to extract bounding boxes was pre-trained on the COCO dataset \cite{coco} which does not contain \texttt{CounterTop} as an object.
PIGLeT-Vis is therefore ultimately limited by the capabilities of its vision backbone.

\section*{Ethics Statement}

While this work does not introduce new data or involve human participants, we use the PIGPeN dataset which contains human-labelled data.
The fine-tuning portion of the dataset was annotated through MTurk by \citet{zellers} and they report following best practices (paying decent wages, providing feedback and using a qualification test) in their data collection. 
We filter and use a subset of PIGPeN and introduce methods to learn the effects of actions in a multimodal setting. We, therefore, believe that our work does not raise any ethical concerns.

\section*{Acknowledgements}

This work was supported in part by the UKRI Centre for Doctoral Training in Natural Language Processing, funded by the UKRI (grant EP/S022481/1) at the University of Edinburgh, School of Informatics and School of Philosophy, Psychology \& Language Sciences and by the UKRI-funded TAS Governance Node (grant number EP/V026607/1).

% Entries for the entire Anthology, followed by custom entries
\bibliography{custom}
\bibliographystyle{acl_natbib}

\appendix
\section{GPT-3 Example of Physical Reasoning}
\label{appendix:gpt3}

\begin{figure}[ht]
    \centering
    \begin{tabular}{ | c |}
    \hline
     \textbf{The weight of the potato is 150 grams.} \\ 
     \textbf{The robot then slices the potato into thin slices.} \\  
     \textbf{The weight of the potato is now} \texttt{75 grams.} \\
     \hline
    \end{tabular}
    \caption{Example of incorrect physical commonsense by an LLM. When predicting what comes after the \textbf{input text}, the large 175 billion parameter GPT-3 \cite{lm} \texttt{predicts} that the weight of the potato halves after a slicing action is taken.}
    \label{fig:gpt3}
\end{figure}

\section{PIGPeN-Vis}
\label{appendix:pigpen}
\begin{figure}
    \includegraphics[width=0.4\textwidth]{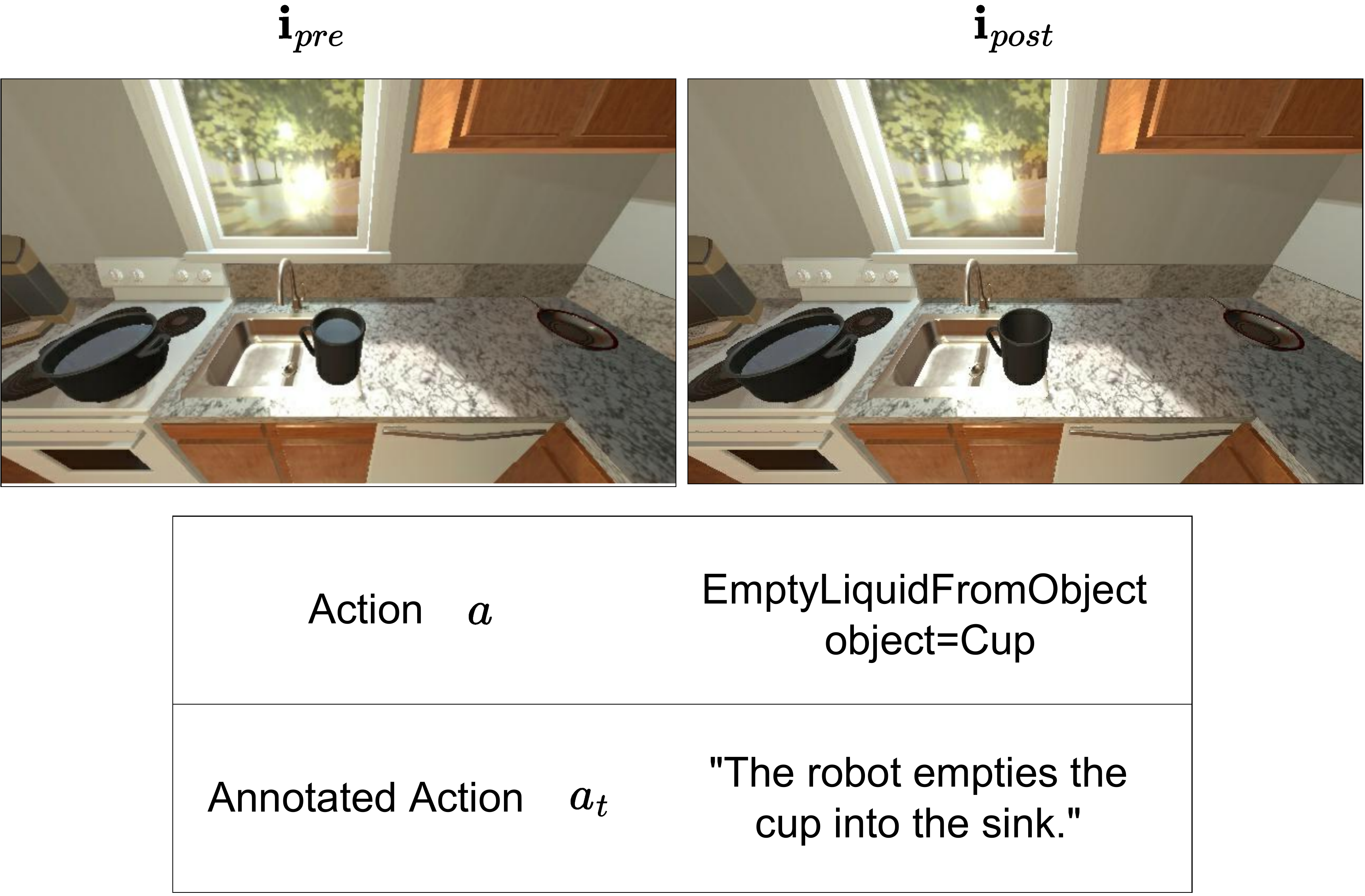}
    \caption{Image pair and actions for a selected PIGPeN example.}
    \label{fig:pigpen_example}
\end{figure}

We select an example from PIGPeN to display in Figure~\ref{fig:pigpen_example} and Table~\ref{table:pigpen_example}.

\begin{table}
\scalebox{0.65}{
\begin{tabular}{l|cc|cc}
& \multicolumn{2}{c}{$pre$} & \multicolumn{2}{c}{$post$} \\
& $\textbf{o}^{cup}_{pre}$ & $\textbf{o}^{faucet}_{pre}$ & $\textbf{o}^{cup}_{post}$ & $\textbf{o}^{faucet}_{post}$ \\
\hline
ObjectName & Cup & Faucet & Cup & Faucet \\
Contained Objects & & & & \\
Is contained in... & & & & \\
Mass & 1 to 2lb & Massless & 1 to 2lb & Massless \\
Size & small & medium & small & medium \\
Temperature & RoomTemp & RoomTemp & RoomTemp & RoomTemp \\
Distance & 1 to 2ft & 3 to 4 ft & 1 to 2ft & 3 to 4 ft \\
Breakable & Yes & No & Yes & No \\
Cookable & No & No & No & No \\
CanBecomeDirty & Yes & No & Yes & No \\
IsBroken & No & No & No & No \\
IsCooked & No & No & No & No \\
IsDirty & No & No & No & No \\
IsFilledWithLiquid & Yes & No & \textbf{No} & No \\
IsOpen & No & No & No & No \\
IsPickedUp & Yes & No & Yes & No \\
IsSliced & No & No & No & No \\
IsToggled & No & No & No & No \\
Moveable & No & No & No & No \\
Openable & No & No & No & No \\
Pickupable & Yes & No & Yes & No \\
CanHoldItems & Yes & No & Yes & No \\
Sliceable & No & No & No & No \\
Toggleable & No & Yes & No & Yes \\
Materials & Ceramic & & Ceramic & \\
\end{tabular}
}
\caption{Attributes for a selected PIGPeN example. The total number of attributes is $38$ as the Materials attribute is a multi-hot encoding.}
\label{table:pigpen_example}
\end{table}

From the original dataset, we remove two attributes (\texttt{isUsedUp} and \texttt{salientMaterials\_Organic}) because they are unchanged in all examples. 
We also remove $3$ actions (\texttt{ThrowObject10}, \texttt{ThrowObject100} and \texttt{ThrowObject1000}) which are all related to throwing an object across a certain distance.
These actions account for only a small subset of the dataset and create inconsistent image pairs due to the agent's momentum being captured in the images.
The angle of the camera changes as a result of \texttt{ThrowObject} and this breaks our assumption that the difference between $\textbf{i}_{pre}$ and $\textbf{i}_{post}$ solely reflects the effects of the action on the environment (and not on the viewer).
We therefore reduce the total number of symbolic attributes per object to $38$ and the number of possible actions to $10$.

\subsection{Attributes}

The following 38 symbolic attributes are used to describe an object in PIGPeN:

\begin{small}
\texttt{
ObjectName, parentReceptacles, receptacleObjectIds, distance, mass,size, ObjectTemperature, breakable, cookable, dirtyable, isBroken, isCooked, isDirty, isFilledWithLiquid, isOpen, isPickedUp, isSliced, isToggled, moveable, openable, pickupable, receptacle, salientMaterials\_Ceramic, salientMaterials\_Fabric, salientMaterials\_Food, salientMaterials\_Glass, salientMaterials\_Leather, salientMaterials\_Metal, salientMaterials\_Paper, salientMaterials\_Plastic, salientMaterials\_Rubber, salientMaterials\_Soap, salientMaterials\_Sponge, salientMaterials\_Stone, salientMaterials\_Wax, salientMaterials\_Wood, sliceable, toggleable
}
\end{small}
\subsection{Filtering Statistics}
\label{sec:filter_stats}

\begin{figure}
    \includegraphics[width=0.4\textwidth]{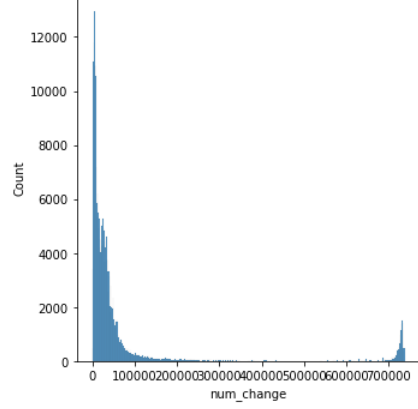}
    \caption{Distribution of the number of pixels changed per image in the PIGPeN dataset.}
    \label{fig:num_change}
\end{figure}

We initially filter the PIGPeN dataset using two main strategies to remove images with too much or too little change between the pre and post images.
In both cases, the goal is to remove pairs of images in which it would be impossible for a vision model to predict what has changed.

Images with too many changes are often images taken from different viewpoints or with different lighting conditions. 
We filter these images by looking at the number of pixels changed between $\textbf{i}_{pre}$ and $\textbf{i}_{post}$.
We show the distribution of the number of pixels changed per image over the training dataset in Figure~\ref{fig:num_change}. 
Using this visualization we can clearly see a small peak at the extreme - where almost all the pixels in $\textbf{i}_{post}$ are different from  $\textbf{i}_{pre}$.
Note that since each image is an RGB image of dimensions $640 \times 385$, the max number of change is $640 \times 385 \times 3 = 739,200$ (we also compare pixels across color channels).
We opt to remove all images with more than $400,000$ changes, which corresponds to around $6.2\%$ of the training dataset.

\begin{figure}
    \includegraphics[width=0.4\textwidth]{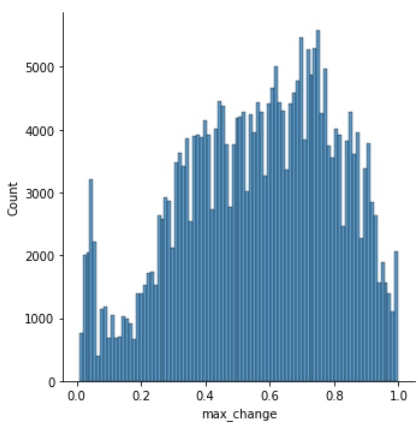}
    \caption{Distribution of the maximum pixel value changed per image in the PIGPeN dataset.}
    \label{fig:max_change}
\end{figure}

Images with too little change could be examples of where the action has no visual outcome and $\textbf{i}_{pre}$ and $\textbf{i}_{post}$ are indistinguishable. 
To filter these images we measure the maximum magnitude of change in each pixel and each color channel between the pairs of images.
We visualize the max change across the training dataset in Figure~\ref{fig:max_change}. 
Here a low values implies almost no salient change, and as max change approaches zero - it becomes unlikely that a human would be able to perceive the difference between the pair of images.
We opt for to keep images with a max change greater than $0.2$ which corresponds to excluding $7.8\%$ of the training dataset.

Filtering on the number of changed pixels lead to the exclusion of around $13.89\%$ of the training dataset.

\subsection{Zero-shot Filtering}
\label{sec:zero_app}

We remove the following 14 objects from both the train and validation ($3,401$ examples total):
\\
\begin{small}
\texttt{HandTowel, Towel, Plunger, Watch, CD, SoapBottle,
Pen, RemoteControl, SoapBar, Box, Bottle, CreditCard,
Statue, KeyChain}
\end{small}

We remove the following 27 action-object pairs from both the train and validation ($3,278$ examples total):
\\
\begin{small}
\texttt{
(CloseObject,Toilet),
(DirtyObject,Pan),
(DirtyObject,Pot),
(EmptyLiquidFromObject,Bottle),
(EmptyLiquidFromObject,Pot),
(OpenObject,Toilet),
(PickupObject,Box),
(PickupObject,CellPhone),
(PickupObject,CreditCard),
(PickupObject,KeyChain),
(PutObject,CD),
(PutObject,CreditCard),
(PutObject,HandTowel),
(PutObject,Laptop),
(PutObject,Lettuce),
(PutObject,Pen),
(PutObject,Plunger),
(PutObject,Pot),
(PutObject,RemoteControl),
(PutObject,SoapBar),
(PutObject,SoapBottle),
(PutObject,Statue),
(PutObject,ToiletPaper),
(PutObject,Towel),
(PutObject,Watch),
(ToggleOff,CellPhone),
(ToggleOff,Television)
}
\end{small}

\section{Code Release and Training}

Our full code, models, and PIGPeN-Vis split can be found at \href{https://github.com/gautierdag/piglet-vis}{github.com/gautierdag/piglet-vis}.
% We also attach an anonymised code release as a \texttt{.zip} file for the reviewers.

\subsection{Additional Training Details}
\label{appendix:training_details}

As previously mentioned, there are a few differences between the original \citet{zellers} model and our implementation of \texttt{base+symbolic}.
We use an off-the-shelf RoBERTa-base \cite{roberta} model instead of a custom GPT2 \cite{radford}.
Additionally, we also reduce the dimensionality of the PIGLeT layers from $h=256$ to $h=64$. 
This shrinks the overall model (excluding the LLM) from $11.9$ million parameters to less than $2$ million parameters during pre-training and improves the overall accuracy by a small margin ($+1.51\%$).
We do not run any other hyper-parameter search throughout our experiments and wherever possible use the same hyper-parameters as PIGLeT.
We also reduce the batch size from $1024$ to $256$ because we use a mix of NVIDIA GTX 1080 and NVIDIA A100 GPUs and wish to keep batch size constant.

The \texttt{+images} models use the extracted representations from a frozen off-the-shelf DETR model ($41.3$ million parameters), however it is ran only once over all images as we cache its predictions.
We do not use the “NO OBJECT” predictions from DETR, and simply pass all 100 bounding boxes representations to the attention mechanism. 
Since we do not have access to the true bounding boxes in PIGPeN, we do not fine-tune DETR and therefore ignore its prediction heads which have also been trained on COCO and mismatch our possible objects.

The \texttt{+symbolic} models use the Symbolic Object Encoder which is an additional $800,000$ parameters on its own.
During fine-tuning all models use a RoBERTA-base model ($+120$ million parameters) in the Action Encoder.
The \texttt{+text-label} model also uses the RoBERTA-base model during pre-training, but again this is frozen and its outputs are cached for the full dataset.

We pre-train each model for $80$ epochs and fine-tune for $60$ epochs.
For all setups, pre-training takes between 1 to 2 hours and fine-tuning takes less than 1 hour on an NVIDIA A100 GPU.
We use the Pytorch implementation of the Adam optimizer \cite{adam} and a learning rate of $10^{-3}$ during pre-training and $10^{-5}$ during fine-tuning.
We use early stopping on the validation loss with a patience of $10$ epochs.
We run each setup over $10$ different seeds ($s\in[1,2,...,10]$ and report the average and standard deviation for each metric.

%   A clear description of the mathematical setting, algorithm, and/or model.
%     Submission of a zip file containing source code, with specification of all dependencies, including external libraries, or a link to such resources (while still anonymized)
%     Description of computing infrastructure used
%     The average runtime for each model or algorithm (e.g., training, inference, etc.), or estimated energy cost
%     Number of parameters in each model
%     Corresponding validation performance for each reported test result
%     Explanation of evaluation metrics used, with links to code

\section{Accuracy Results}
\label{appendix:results}

\begin{table}[]
\scalebox{0.7}{
\begin{tabular}{lllr}
\hline
 & \multicolumn{3}{c}{\textbf{Overall Accuracy ($\% \pm \sigma$)}} \\
 & PIGPeN & PIGPeN-Vis & \multicolumn{1}{l}{$\Delta$} \\
\texttt{base} & $29.18\pm0.34$ & $21.23\pm0.72$ & $-7.95\%$ \\
\hline 
\texttt{base+symbolic (PIGLeT)} & $86.39\pm0.79$ & $85.03\pm0.45$ & $-1.36\%$ \\
\texttt{base+symbolic+images} & $87.45\pm0.66$ & $86.01\pm0.89$ & $-1.44\%$ \\
\hline 
\texttt{base+images (PIGLet-Vis)} & $49.13\pm1.53$ & $45.47\pm1.50$ & $-3.66\%$ \\
\texttt{base+images+text-labels} & $51.28\pm1.68$ & $47.55\pm2.10$ & $-3.73\%$ \\
\hline 
\end{tabular}
}
\caption{Overall Accuracies comparing full PIGPeN with the PIGPeN-Vis split across 10 seeds.}
\label{tab:results_pigpen}
\end{table}

\subsection{Comparing PIGPeN and PIGPeN-Vis}
Table~\ref{tab:results_pigpen} compares the overall accuracy on the original PIGPeN dataset with our proposed PIGPeN-Vis split. 
We find that our PIGPeN-Vis split is consistently harder to solve than the original PIGPeN dataset.
We explain the increased accuracy in the original dataset with the fact that some of the filtered out actions (see Appendix~\ref{appendix:pigpen}) are easy to solve from knowing the object name and action: e.g., most of the images we exclude due to little salient changes are appliances like stoves being turned on or off. 
However, it is easy for a model to predict the post-condition attributes of a stove, which are mostly static, across all examples given an action such as \texttt{ToggleOn}, which always has the same effect. 
% Through this comparison, we also find that our model is capable of inferring and ignoring uninformative or unimportant visual inputs.

\subsection{Complete Accuracy Results on PIGPeN-Vis}
Table~\ref{tab:results_val} shows the overall accuracies for both the test and validation sets.
The full accuracy results for all actions in Table~\ref{table:act_accs} and for all attributes in Table~\ref{table:att_accs}.

\begin{table}[t]
\centering
\scalebox{0.7}{
\begin{tabular}{lll}
&\multicolumn{2}{c}{\textbf{Overall Accuracy} ($\% \pm \sigma$)}\\
&validation & test\\
\hline 
\texttt{base}& $23.85\pm0.95$& $21.23\pm0.72$\\
\hline 
\texttt{base+symbolic} (PIGLeT)& $88.08\pm0.50$& $85.03\pm0.45$\\
\texttt{base+symbolic+images}&$\bm{89.49\pm0.82}$& $\bm{86.01\pm0.89}$\\
\hline 
\texttt{base+images}&$50.73\pm2.97$& $45.47\pm1.50$\\
\texttt{base+images+text-labels}&$53.33\pm3.15$& $47.55\pm2.10$\\
\hline
\end{tabular}
}
\caption{Validation and test overall accuracies. Note the zero-shot accuracy is not calculated on the validation set since there are no unseen examples in the validation set to prevent leakage.}
\label{tab:results_val}
\end{table}

\begin{table*}[t]
\centering
\scalebox{0.7}{
\begin{tabular}{lccccc}
& \multicolumn{5}{c}{\textbf{Action Accuracy} ($\%\pm\sigma$)}\\
\hline
&\texttt{Close}&\texttt{Dirty}&\texttt{EmptyLiquid}&\texttt{HeatUpPan}&\texttt{Open}\\
\hline
\texttt{base}& $13.20\pm1.06$& $17.71\pm1.20$& $24.75\pm5.75$& $36.33\pm4.14$& $8.33\pm1.84$\\
\hline
\texttt{base+symbolic}& $85.98\pm1.77$& $94.00\pm3.42$& $99.34\pm1.15$& $100.00\pm0.00$& $85.73\pm0.99$\\
\texttt{base+symbolic+images}& $86.80\pm3.29$& $90.29\pm5.90$& $99.02\pm2.07$& $99.17\pm1.62$& $88.75\pm3.02$\\
\hline
\texttt{base+images}& $27.42\pm3.71$& $58.57\pm2.78$& $69.34\pm4.17$& $68.67\pm3.75$& $20.83\pm4.63$\\
\texttt{base+images+text-labels}& $28.87\pm3.19$& $57.71\pm3.24$& $70.16\pm3.17$& $74.00\pm3.16$& $22.92\pm5.79$\\
\hline
&\texttt{Pickup}&\texttt{Put}&\texttt{Slice}&\texttt{ToggleOff}&\texttt{ToggleOn}\\
\hline
\texttt{base}& $10.96\pm1.92$& $27.95\pm1.19$& $22.13\pm0.86$& $30.83\pm3.39$& $27.38\pm2.57$\\
\hline
\texttt{base+symbolic}& $80.48\pm2.88$& $58.39\pm1.94$& $75.41\pm1.89$& $99.40\pm0.84$& $96.90\pm1.61$\\
\texttt{base+symbolic+images}& $86.14\pm2.56$& $57.59\pm2.31$& $81.31\pm3.96$& $99.05\pm0.75$& $92.86\pm5.14$\\
\hline
\texttt{base+images}& $33.49\pm3.45$& $34.91\pm2.43$& $41.64\pm3.80$& $71.43\pm2.75$& $70.24\pm16.00$\\
\texttt{base+images+text-labels}& $40.12\pm2.61$& $38.30\pm3.11$& $45.57\pm3.85$& $69.05\pm5.81$& $67.14\pm16.53$\\
\end{tabular}
}
\caption{Full accuracy results table including the standard deviation over 10 seeds for all actions and setups.}
\label{table:act_accs}
\end{table*}

\begin{table*}[t]
\centering
\scalebox{0.5}{
\begin{tabular}{lcccccccccc}
& \multicolumn{10}{c}{\textbf{Attribute Accuracy} ($\%\pm\sigma$)}\\
\hline
&\texttt{Name}&\texttt{Temperature}&\texttt{attribute}&\texttt{breakable}&\texttt{cookable}&\texttt{dirtyable}&\texttt{distance}&\texttt{isBroken}&\texttt{isCooked}&\texttt{isDirty}\\
\hline
\texttt{base}& $99.66\pm0.07$& $95.91\pm0.41$& $96.12\pm0.07$& $91.46\pm0.36$& $99.95\pm0.07$& $99.95\pm0.10$& $51.01\pm0.93$& $99.86\pm0.00$& $98.60\pm0.06$& $97.93\pm0.19$\\
\texttt{base+symbolic}& $99.64\pm0.12$& $99.85\pm0.04$& $99.48\pm0.03$& $99.84\pm0.09$& $100.00\pm0.00$& $100.00\pm0.00$& $95.13\pm0.35$& $100.00\pm0.00$& $99.85\pm0.04$& $99.71\pm0.14$\\
\texttt{base+symbolic+images}& $99.62\pm0.09$& $99.59\pm0.27$& $99.48\pm0.04$& $99.78\pm0.10$& $100.00\pm0.00$& $99.97\pm0.09$& $96.13\pm0.40$& $100.00\pm0.00$& $99.85\pm0.04$& $99.52\pm0.32$\\
\texttt{base+images}& $97.34\pm0.65$& $96.28\pm0.74$& $97.25\pm0.13$& $92.63\pm0.75$& $99.91\pm0.10$& $99.62\pm0.20$& $76.90\pm1.05$& $99.85\pm0.05$& $98.68\pm0.19$& $97.87\pm0.34$\\
\texttt{base+images+text-labels}& $98.44\pm0.35$& $96.05\pm1.23$& $97.46\pm0.13$& $93.19\pm0.31$& $99.96\pm0.09$& $99.93\pm0.10$& $78.56\pm1.16$& $99.84\pm0.09$& $98.19\pm0.84$& $97.78\pm0.24$\\
\hline
&\texttt{isFilledWithLiquid}&\texttt{isOpen}&\texttt{isPickedUp}&\texttt{isSliced}&\texttt{isToggled}&\texttt{mass}&\texttt{moveable}&\texttt{openable}&\texttt{parentReceptacles}&\texttt{pickupable}\\
\hline
\texttt{base}& $96.79\pm0.50$& $98.84\pm0.23$& $94.83\pm0.82$& $97.99\pm0.09$& $98.36\pm0.23$& $96.51\pm0.15$& $99.90\pm0.09$& $99.97\pm0.06$& $87.44\pm0.42$& $99.84\pm0.09$\\
\texttt{base+symbolic}& $99.93\pm0.12$& $98.95\pm0.09$& $99.27\pm0.31$& $100.00\pm0.00$& $99.88\pm0.12$& $99.33\pm0.14$& $99.99\pm0.04$& $99.97\pm0.06$& $97.78\pm0.47$& $99.90\pm0.11$\\
\texttt{base+symbolic+images}& $99.84\pm0.19$& $98.67\pm0.38$& $98.96\pm0.31$& $99.97\pm0.06$& $99.74\pm0.30$& $99.59\pm0.09$& $100.00\pm0.00$& $99.99\pm0.04$& $97.26\pm0.44$& $99.88\pm0.10$\\
\texttt{base+images}& $96.88\pm0.55$& $98.81\pm0.97$& $97.43\pm0.37$& $98.28\pm0.30$& $97.92\pm0.83$& $96.41\pm0.41$& $99.79\pm0.21$& $99.74\pm0.20$& $91.05\pm0.77$& $99.59\pm0.17$\\
\texttt{base+images+text-labels}& $97.25\pm0.45$& $98.11\pm1.14$& $97.54\pm0.53$& $98.34\pm0.29$& $98.06\pm0.55$& $96.74\pm0.24$& $99.89\pm0.09$& $99.95\pm0.10$& $92.49\pm0.69$& $99.70\pm0.09$\\
\hline
&\texttt{receptacleIds}&\texttt{receptacle}&\texttt{Ceramic}&\texttt{Fabric}&\texttt{Food}&\texttt{Glass}&\texttt{Leather}&\texttt{Metal}&\texttt{Paper}&\texttt{Plastic}\\
\hline
\texttt{base}& $84.20\pm0.61$& $99.85\pm0.10$& $98.26\pm0.17$& $99.55\pm0.07$& $99.99\pm0.04$& $98.91\pm0.13$& $99.89\pm0.06$& $98.69\pm0.15$& $99.73\pm0.00$& $98.30\pm0.10$\\
\texttt{base+symbolic}& $96.36\pm0.18$& $99.90\pm0.09$& $100.00\pm0.00$& $99.96\pm0.07$& $100.00\pm0.00$& $99.99\pm0.04$& $100.00\pm0.00$& $99.99\pm0.04$& $100.00\pm0.00$& $99.97\pm0.06$\\
\texttt{base+symbolic+images}& $96.13\pm0.30$& $99.92\pm0.10$& $99.99\pm0.04$& $99.85\pm0.10$& $99.99\pm0.04$& $99.97\pm0.06$& $100.00\pm0.00$& $100.00\pm0.00$& $100.00\pm0.00$& $99.96\pm0.07$\\
\texttt{base+images}& $82.87\pm0.55$& $99.47\pm0.21$& $99.03\pm0.22$& $99.50\pm0.19$& $99.92\pm0.10$& $99.16\pm0.21$& $99.97\pm0.06$& $98.31\pm0.37$& $99.67\pm0.21$& $98.83\pm0.31$\\
\texttt{base+images+text-labels}& $83.91\pm0.56$& $99.69\pm0.11$& $99.36\pm0.19$& $99.44\pm0.12$& $99.96\pm0.09$& $99.37\pm0.24$& $99.95\pm0.10$& $98.69\pm0.30$& $99.56\pm0.19$& $99.08\pm0.20$\\
\hline
&\texttt{Rubber}&\texttt{Soap}&\texttt{Sponge}&\texttt{Stone}&\texttt{Wax}&\texttt{Wood}&\texttt{size}&\texttt{sliceable}&\texttt{toggleable}\\
\hline
\texttt{base}& $100.00\pm0.00$& $99.99\pm0.04$& $100.00\pm0.00$& $99.34\pm0.09$& $100.00\pm0.00$& $99.51\pm0.16$& $73.78\pm0.29$& $98.02\pm0.12$& $99.95\pm0.07$\\
\texttt{base+symbolic}& $100.00\pm0.00$& $100.00\pm0.00$& $100.00\pm0.00$& $99.99\pm0.04$& $100.00\pm0.00$& $99.99\pm0.04$& $94.98\pm0.19$& $100.00\pm0.00$& $99.99\pm0.04$\\
\texttt{base+symbolic+images}& $99.97\pm0.06$& $99.99\pm0.04$& $100.00\pm0.00$& $99.99\pm0.04$& $100.00\pm0.00$& $100.00\pm0.00$& $96.35\pm0.20$& $99.99\pm0.04$& $99.96\pm0.09$\\
\texttt{base+images}& $99.88\pm0.08$& $99.89\pm0.11$& $99.92\pm0.10$& $99.48\pm0.14$& $99.92\pm0.10$& $99.25\pm0.22$& $87.03\pm1.15$& $98.32\pm0.32$& $99.81\pm0.17$\\
\texttt{base+images+text-labels}& $99.85\pm0.08$& $99.92\pm0.10$& $99.88\pm0.14$& $99.60\pm0.19$& $99.95\pm0.07$& $99.37\pm0.22$& $87.89\pm1.11$& $98.32\pm0.36$& $99.95\pm0.07$\\
\end{tabular}
}
\caption{Full accuracy results table including the standard deviation over 10 seeds for all attributes and setups.}
\label{table:att_accs}
\end{table*}

\section{Additional Attention Maps}

\begin{figure*}
     \centering
     \begin{subfigure}[b]{0.3\textwidth}
         \centering
         \includegraphics[width=\textwidth]{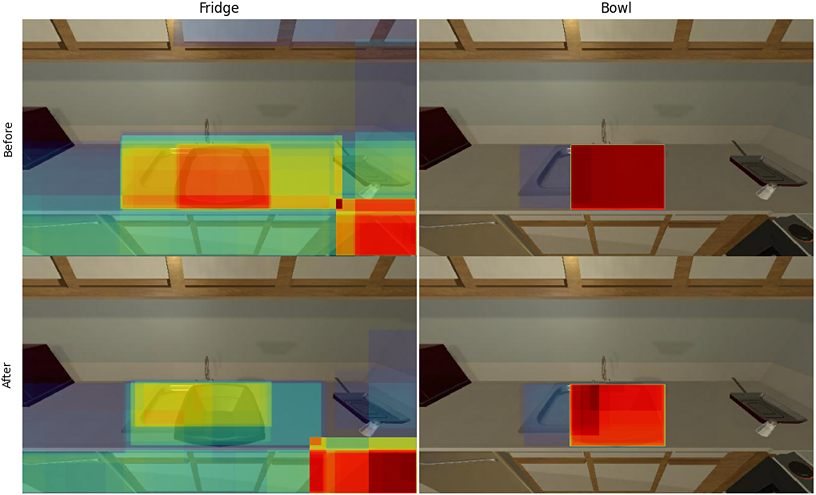}
         \caption{\texttt{base+images}}
         \label{fig:empty_i}
     \end{subfigure}
     \hfill
     \begin{subfigure}[b]{0.3\textwidth}
         \centering
         \includegraphics[width=\textwidth]{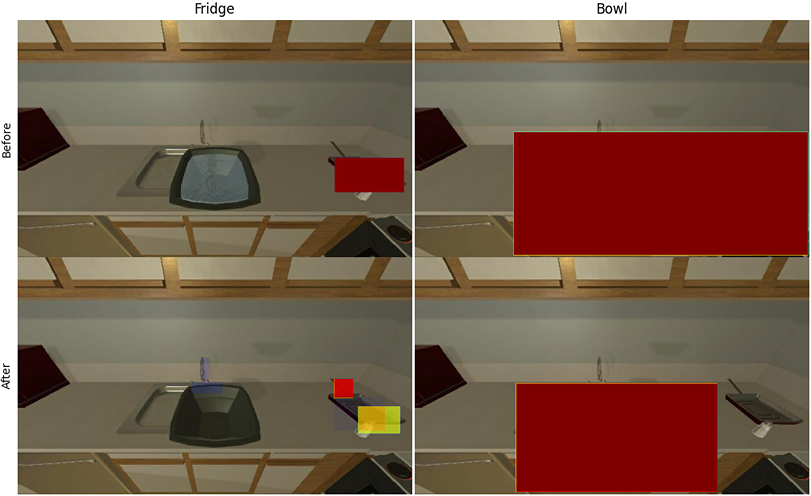}
         \caption{\texttt{base+symbolic+images}}
         \label{fig:empty_s}
     \end{subfigure}
     \hfill
     \begin{subfigure}[b]{0.3\textwidth}
         \centering
         \includegraphics[width=\textwidth]{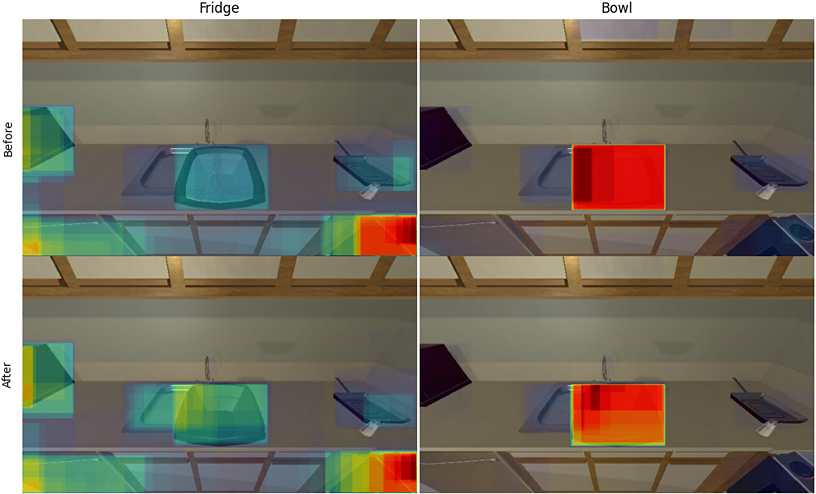}
         \caption{\texttt{base+images+text-labels}}
         \label{fig:empty_t}
     \end{subfigure}
        \caption{Attention maps for the effects of the \texttt{EmptyLiquid} action on \texttt{Bowl} with objects \texttt{Fridge} and \texttt{Bowl}. The top row of each grid maps to the before environment and the bottom row maps to the after environment. The columns map to each respective object. The \texttt{Fridge} object appears in the lower left of the image, and is only correctly identified by \texttt{base+images+text-labels}, even though the model does place more weight to the bounding box of the stove (lower right). }
        \label{fig:empty_maps}
\end{figure*}

\begin{figure*}
     \centering
     \begin{subfigure}[b]{0.3\textwidth}
         \centering
         \includegraphics[width=\textwidth]{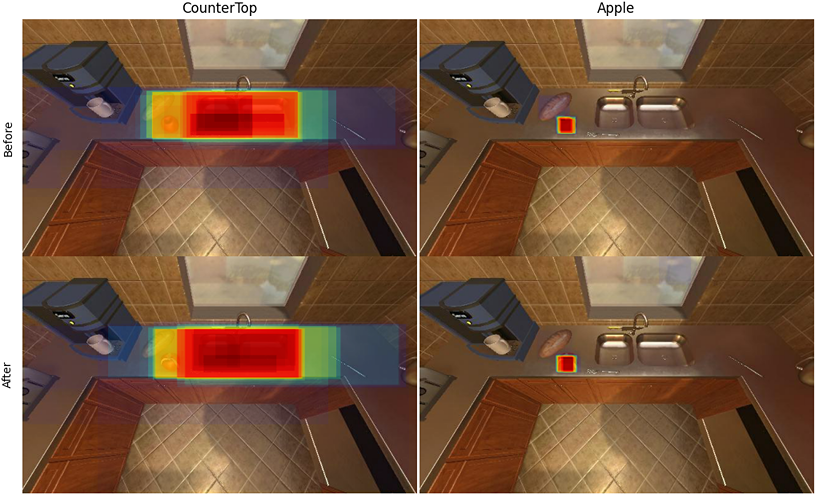}
         \caption{\texttt{base+images}}
         \label{fig:slice_i}
     \end{subfigure}
     \hfill
     \begin{subfigure}[b]{0.3\textwidth}
         \centering
         \includegraphics[width=\textwidth]{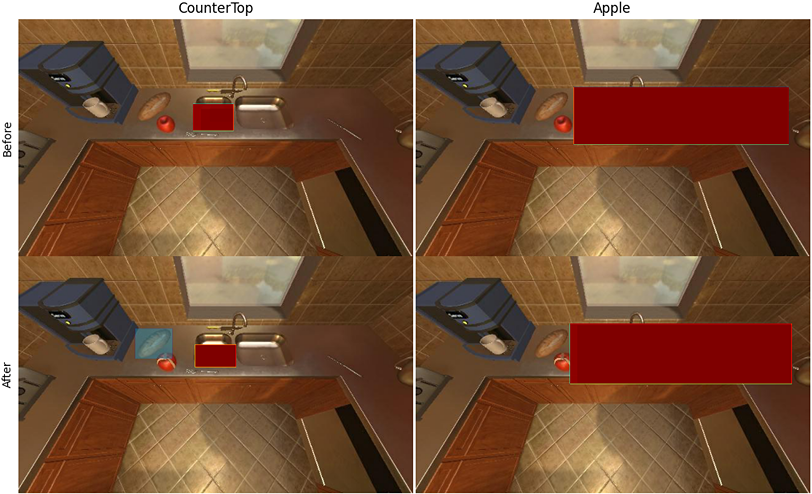}
         \caption{\texttt{base+symbolic+images}}
         \label{fig:slice_s}
     \end{subfigure}
     \hfill
     \begin{subfigure}[b]{0.3\textwidth}
         \centering
         \includegraphics[width=\textwidth]{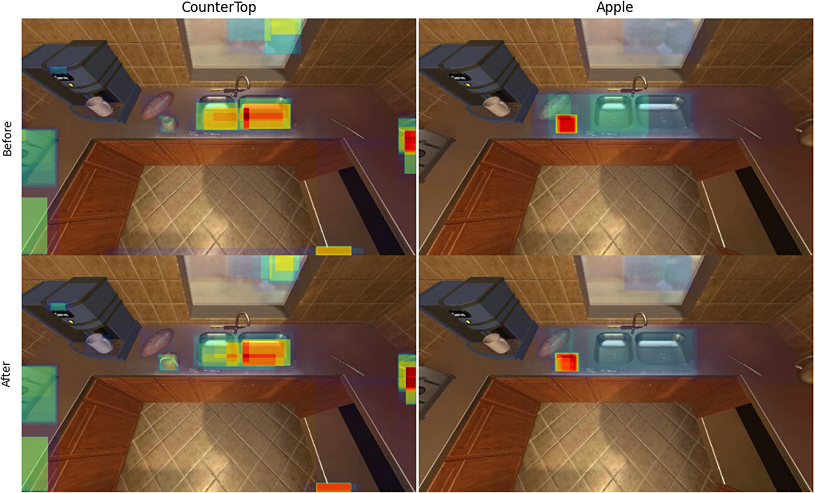}
         \caption{\texttt{base+images+text-labels}}
         \label{fig:slice_t}
     \end{subfigure}
        \caption{Attention maps for the effects of the \texttt{Slice} action on \texttt{Apple} with objects \texttt{CounterTop} and \texttt{Apple}. The top row of each grid maps to the before environment and the bottom row maps to the after environment. The columns map to each respective object.}
        \label{fig:slice_maps}
\end{figure*}

\begin{figure*}
     \centering
     \begin{subfigure}[b]{0.3\textwidth}
         \centering
         \includegraphics[width=\textwidth]{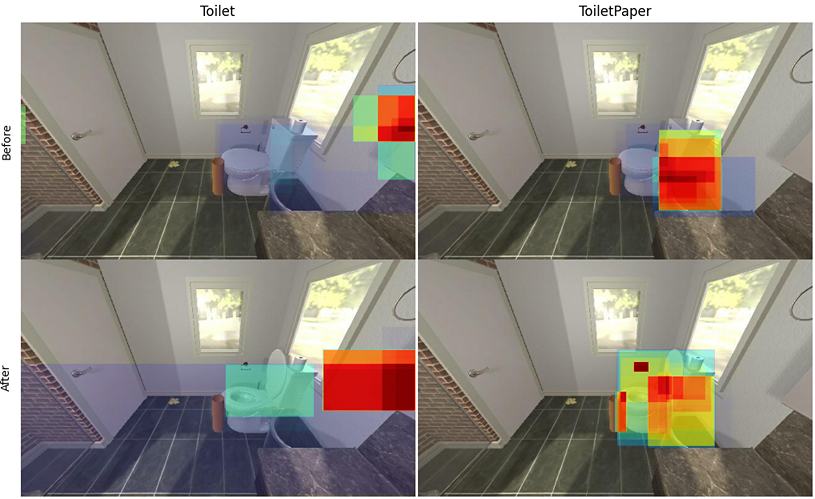}
         \caption{\texttt{base+images}}
         \label{fig:open_i}
     \end{subfigure}
     \hfill
     \begin{subfigure}[b]{0.3\textwidth}
         \centering
         \includegraphics[width=\textwidth]{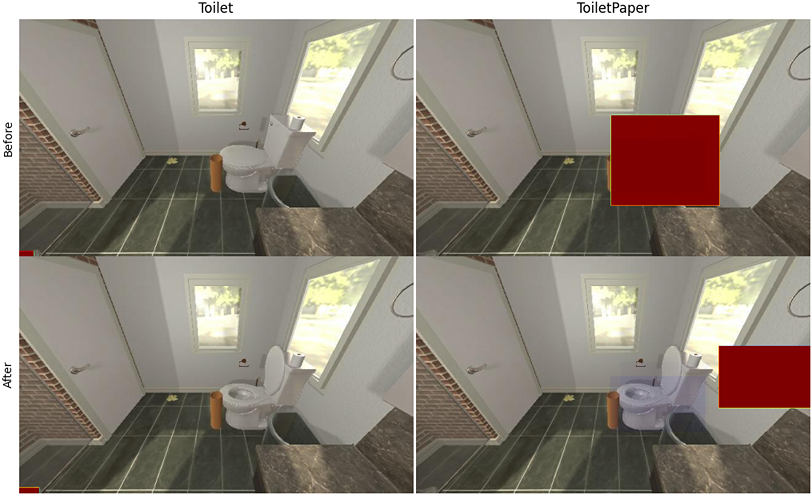}
         \caption{\texttt{base+symbolic+images}}
         \label{fig:open_s}
     \end{subfigure}
     \hfill
     \begin{subfigure}[b]{0.3\textwidth}
         \centering
         \includegraphics[width=\textwidth]{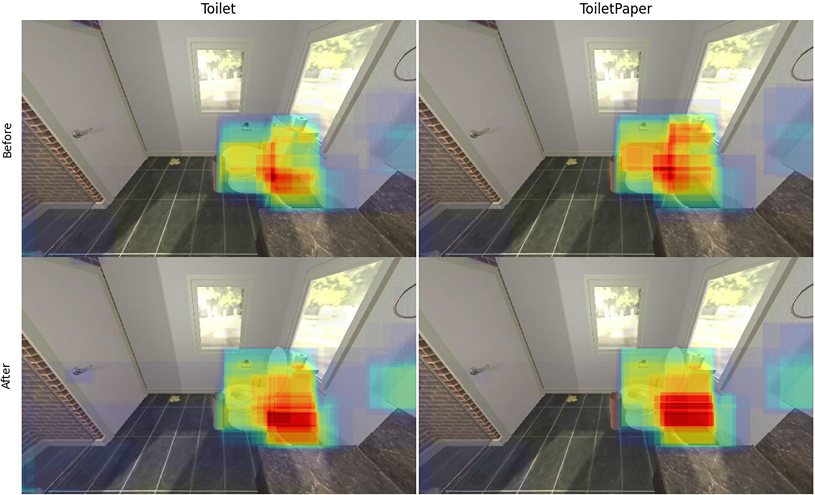}
         \caption{\texttt{base+images+text-labels}}
         \label{fig:open_t}
     \end{subfigure}
        \caption{Attention maps for the effects of the \texttt{Open} action on \texttt{Toilet} with objects \texttt{Toilet} and \texttt{ToiletPaper}. The top row of each grid maps to the before environment and the bottom row maps to the after environment. The columns map to each respective object. This particular example is an unseen combination of action and object that has been excluded from the training and validation set.}
        \label{fig:open_maps}
\end{figure*}

We plot additional attention visualizations for all three image models \texttt{base+images}, \texttt{base+symbolic+images}, and \texttt{base+images+text-labels} in Figures~\ref{fig:empty_maps}, Figures~\ref{fig:slice_maps}, and Figures~\ref{fig:open_maps}.
Since the DETR object detector remains frozen, all models have access to the same bounding boxes and bounding box representations.
Qualitatively, we find that the attention weights of \texttt{base+images} and \texttt{base+images+text-labels} both learn to map to globally relevant bounding boxes given an objects.
We also find the attention maps in \texttt{base+images+text-labels} to be less confident overall than \texttt{base+images}, likely due to the noise introduced by the semantic text inputs. 
As a result, \texttt{base+images+text-labels} makes less mistakes by not focusing too much attention to the wrong bounding box.

On the other hand, \texttt{base+symbolic+images} focuses on seemingly random bounding boxes.
Since \texttt{base+symbolic+images} already receives the full representation of each objects, it does not learn to complement the object's representation with accurate visual information. 
While \texttt{base+symbolic+images} extracts $1\%$ of additional overall accuracy from image inputs when compared to \texttt{base+symbolic}, it does so by falling back to vision for visually salient actions such as \texttt{Pickup}.
\texttt{base+symbolic+images} focuses only a narrow set bounding boxes with overconfidence with no regard for whether or not the bounding box relates to the object. 
We posit that the model might use vision to better estimate more difficult attributes to predict such as \texttt{distance} in some contexts.
Note \texttt{Pickup} is a salient action because when the agent in the environment picks an object up, the object is placed directly in the middle of its field of vision (as if the agent were holding the object in front of it).

\begin{figure*}
     \centering
     \begin{subfigure}[b]{0.3\textwidth}
         \centering
         \includegraphics[width=\textwidth]{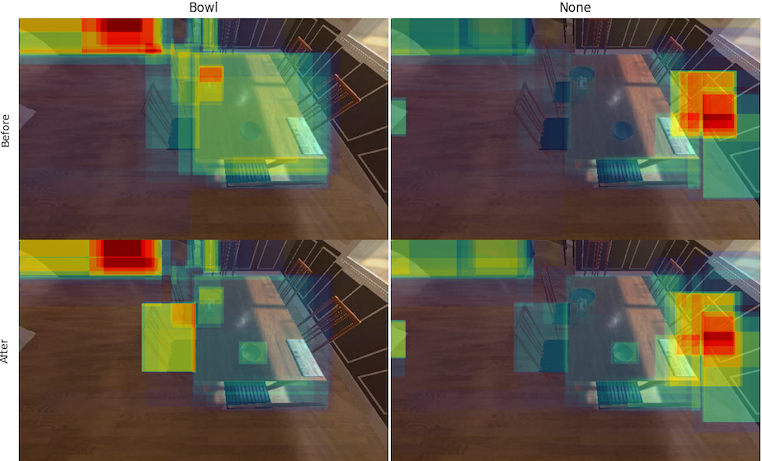}
         \caption{\texttt{base+images}}
         \label{fig:dirty_i}
     \end{subfigure}
     \hfill
     \begin{subfigure}[b]{0.3\textwidth}
         \centering
         \includegraphics[width=\textwidth]{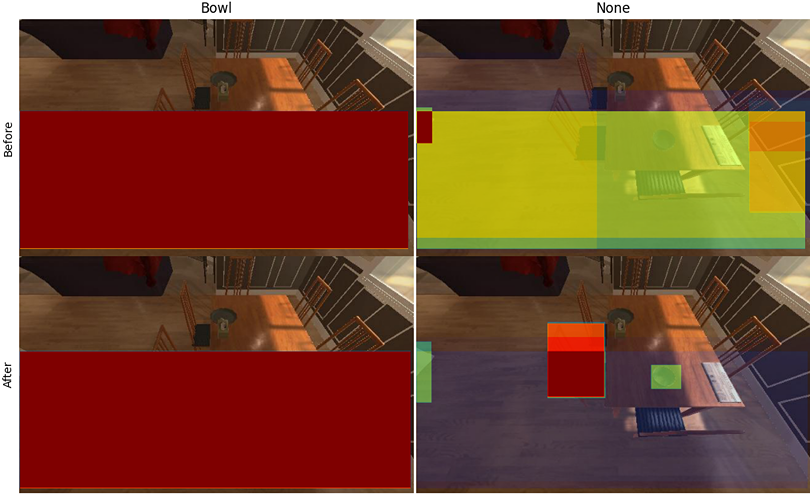}
         \caption{\texttt{base+symbolic+images}}
         \label{fig:dirty_s}
     \end{subfigure}
     \hfill
     \begin{subfigure}[b]{0.3\textwidth}
         \centering
         \includegraphics[width=\textwidth]{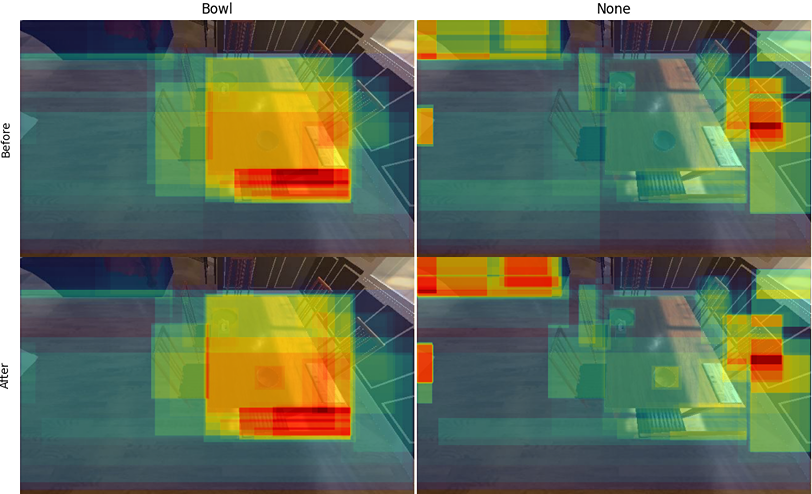}
         \caption{\texttt{base+images+text-labels}}
         \label{fig:dirty_t}
     \end{subfigure}
        \caption{Attention maps for the effects of the \texttt{Dirty} action on \texttt{Bowl} with objects \texttt{Bowl} and \texttt{None}. The top row of each grid maps to the before environment and the bottom row maps to the after environment. The columns map to each respective object. \texttt{None} can be an object in PIGPeN, but we do not predict its attributes and exclude it in all model predictions.}
        \label{fig:dirty_maps}
\end{figure*}

\end{document}